\documentclass[11pt]{article}
\usepackage{acl2016}
\usepackage{times}
\usepackage{url}
\usepackage{latexsym}

\usepackage{algorithm}
\usepackage{algorithmicx}
\usepackage{algpseudocode}

\usepackage{multirow}
\usepackage{graphicx}
\usepackage{amsmath}
\usepackage{mathrsfs}
\usepackage{amssymb}
\usepackage{amsfonts}
\usepackage{color}

\aclfinalcopy 

\title{Bridging Neural Machine Translation and Bilingual Dictionaries}

\author{Jiajun Zhang$^{\dagger}$ \and Chengqing Zong$^{\dagger}${}$^{\ddagger}$\\
$^{\dagger}$University of Chinese Academy of Sciences, Beijing, China\\
National Laboratory of Pattern Recognition, CASIA, Beijing, China\\
$^{\ddagger}$CAS Center for Excellence in Brain Science and Intelligence Technology, Shanghai, China\\
  {\tt \{jjzhang,cqzong\}@nlpr.ia.ac.cn}}

\date{}

\begin{document}
\maketitle
\begin{abstract}
Neural Machine Translation (NMT) has become the new state-of-the-art in several language pairs. However, it remains a challenging problem how to integrate NMT with a bilingual dictionary which mainly contains words rarely or never seen in the bilingual training data. In this paper, we propose two methods to bridge NMT and the bilingual dictionaries. The core idea behind is to design novel models that transform the bilingual dictionaries into adequate sentence pairs, so that NMT can distil latent bilingual mappings from the ample and repetitive phenomena. One method leverages a mixed word/character model and the other attempts at synthesizing parallel sentences guaranteeing massive occurrence of the translation lexicon. Extensive experiments demonstrate that the proposed methods can remarkably improve the translation quality, and most of the rare words in the test sentences can obtain correct translations if they are covered by the dictionary.

\end{abstract}

\section{Introduction}

\begin{figure*}
\centering
\includegraphics[scale=0.5]{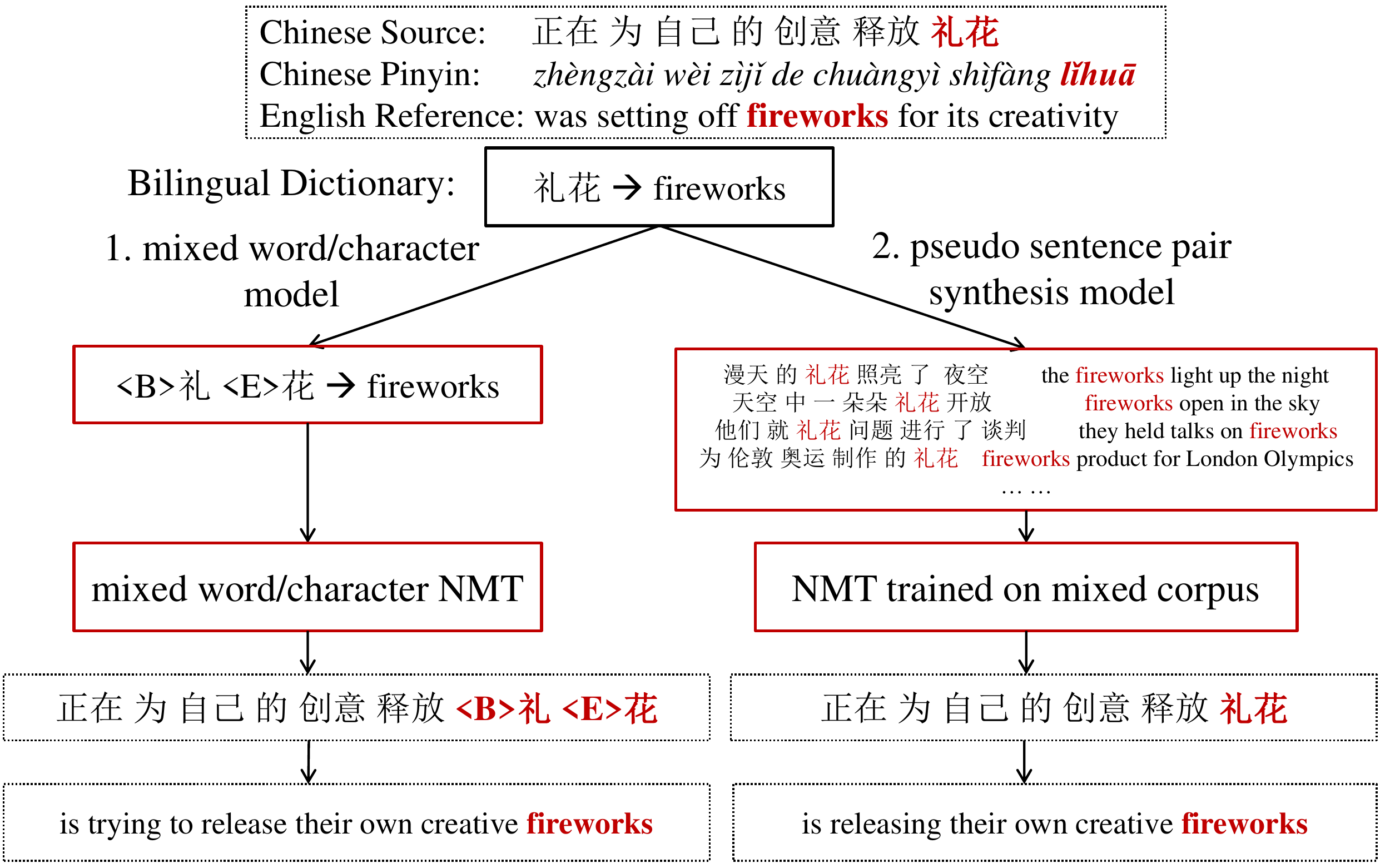}
\caption{The framework of our proposed methods.}
\label{Fig.1}
\end{figure*}

Due to its superior ability in modelling the end-to-end translation process, neural machine translation (NMT), recently proposed by \cite{kalchbrenner:2013,cho:2014,sutskever:2014}, has become the novel paradigm and achieved the new state-of-the-art translation performance for several language pairs, such as English-to-French, English-to-German and Chinese-to-English \cite{sutskever:2014,bahdanau:2014,luong:2015a,sennrich:2015a,wu:2016}.

Typically, NMT adopts the encoder-decoder architecture which consists of two recurrent neural networks. The encoder network models the semantics of the source sentence and transforms the source sentence into the context vector representation, from which the decoder network generates the target translation word by word. 

One important feature of NMT is that each word in the vocabulary is mapped into a low-dimensional real-valued vector (word embedding). The use of continuous representations enables NMT to learn latent bilingual mappings for accurate translation and explore the statistical similarity between words (e.g. {\em desk} and {\em table}) as well. As a disadvantage of the statistical models, NMT can learn good word embeddings and accurate bilingual mappings only when the words occur frequently in the parallel sentence pairs. However, low-frequency words are ubiquitous, especially when the training data is not enough (e.g. low-resource language pairs). Fortunately, in many language pairs and domains, we have handmade bilingual dictionaries which mainly contain words rarely or never seen in the training corpus. Therefore, it remains a big challenge how to bridge NMT and the bilingual dictionaries.

Recently, \newcite{arthur:2016} attempt at incorporating discrete translation lexicons into NMT. The main idea of their method is leveraging the discrete translation lexicons to positively influence the probability distribution of the output words in the NMT softmax layer. However, their approach only addresses the translation lexicons which are in the restricted vocabulary {\footnote{NMT usually keeps only the words whose occurrence is more than a threshold (e.g. 10), since very rare words can not yield good embeddings and large vocabulary leads to high computational complexity.}} of NMT. The out-of-vocabulary (OOV) words are out of their consideration.

In this paper, we aim at making full use of all the bilingual dictionaries, especially the ones covering the rare or OOV words. Our basic idea is to transform the low-frequency word pair in bilingual dictionaries into adequate sequence pairs which guarantee the frequent occurrence of the word pair, so that NMT can learn translation mappings between the source word and the target word.

To achieve this goal, we propose two methods, as shown in Fig. 1. In the test sentence, the Chinese word $l\check{i}hu\bar{a}$ appears only once in our training data and the baseline NMT cannot correctly translate this word. Fortunately, our bilingual dictionary contains this translation lexicon. Our first method extends the mixed word/character model proposed by \newcite{wu:2016} to re-label the rare words in both of the dictionary and training data with character sequences in which characters are now frequent and the character translation mappings can be learnt by NMT. Instead of backing off words into characters, our second method is well designed to synthesize adequate pseudo sentence pairs containing the translation lexicon, allowing NMT to learn the word translation mappings.

We make the following contributions in this paper:

\begin{itemize}

\item We propose a low-frequency to high-frequency framework to bridge NMT and the bilingual dictionaries.

\item We propose and investigate two methods to utilize the bilingual dictionaries. One extends the mixed word/character model and the other designs a pseudo sentence pair synthesis model.

\item The extensive experiments on Chinese-to-English translation show that our proposed methods significantly outperform the strong attention-based NMT. We further find that most of rare words can be correctly translated, as long as they are covered by the bilingual dictionary.

\end{itemize}

\section{Neural Machine Translation}

Our framework bridging NMT and the discrete bilingual dictionaries can be applied in any neural machine translation model. Without loss of generality, we use the attention-based NMT proposed by \cite{luong:2015a}, which utilizes stacked Long-Short Term Memory (LSTM, \cite{hochreiter:1997}) layers for both encoder and decoder as illustrated in Fig. 2. 

The encoder-decoder NMT first encodes the source sentence $X=(x_1,x_2,\cdots,x_{T_x})$ into a sequence of context vectors $C=(\mathbf{h}_1,\mathbf{h}_2,\cdots,\mathbf{h}_{T_x})$ whose size varies with respect to the source sentence length. Then, the encoder-decoder NMT decodes from the context vectors $C$ and generates target translation $Y=(y_1,y_2,\cdots,y_{T_y})$ one word each time by maximizing the probability of $p(y_i|y_{<i},C)$. Note that $x_j$ ($y_i$) is word embedding corresponding to the $j_{th}$ ($i_{th}$) word in the source (target) sentence. Next, we briefly review the encoder introducing how to obtain $C$ and the decoder addressing how to calculate $p(y_i|y_{<i},C)$.

\begin{figure}
\centering
\includegraphics[scale=0.55]{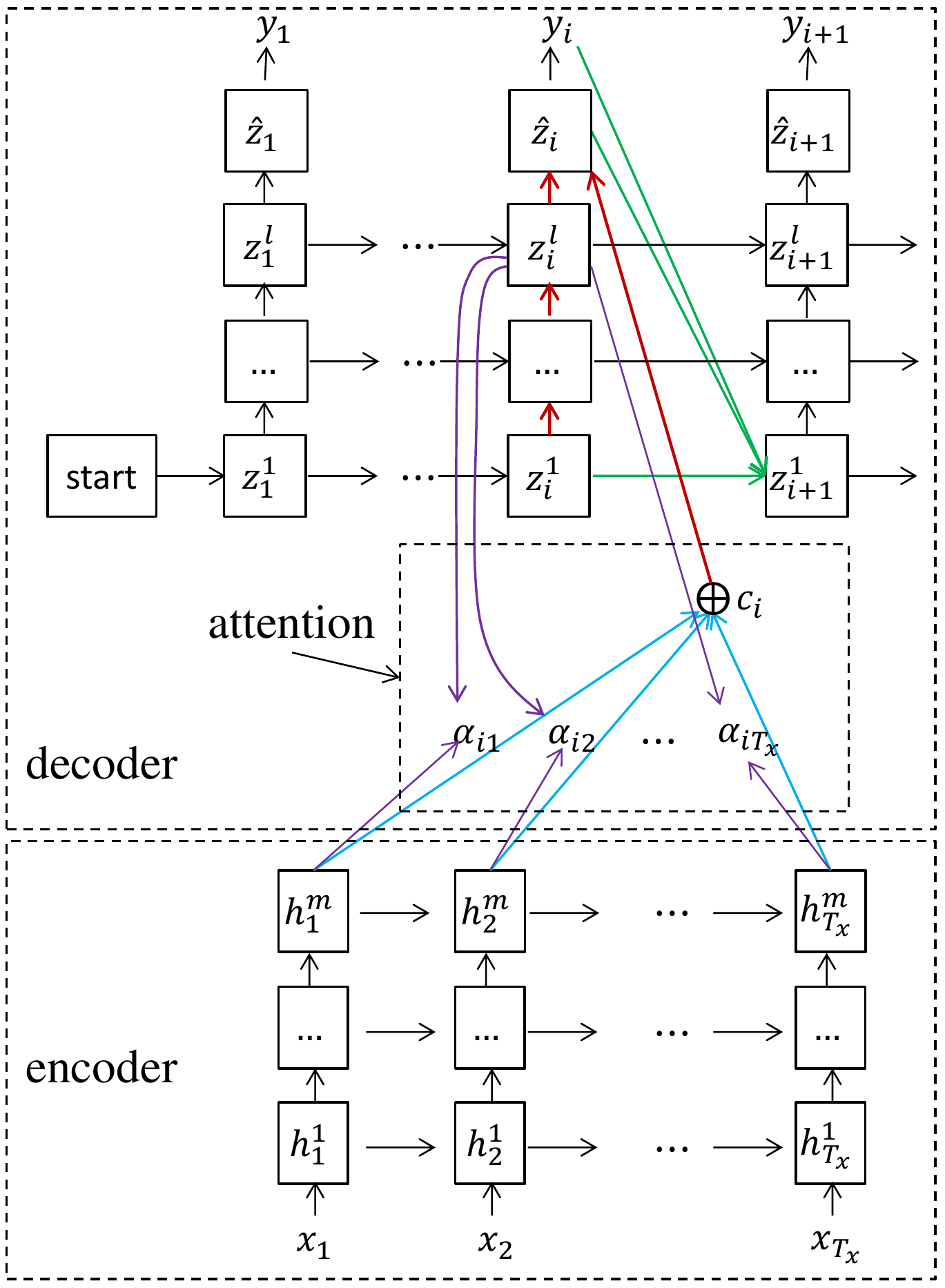}
\caption{The architecture of the attention-based NMT which has $m$ stacked LSTM layers for encoder and $l$ stacked LSTM layers for decoder.}
\label{Fig.2}
\end{figure}

{\bf Encoder}: The context vectors $C=(\mathbf{h}^m_1,\mathbf{h}^m_2,\cdots,\mathbf{h}^m_{T_x})$ are generated by the encoder using $m$ stacked LSTM layers. $\mathbf{h}^k_j$ is calculated as follows:

\begin{equation}
\mathbf{h}^k_j = LSTM(\mathbf{h}^k_{j-1}, \mathbf{h}^{k-1}_j)
\end{equation}

Where $\mathbf{h}^{k-1}_j=x_j$ if $k=1$.

{\bf Decoder}: The conditional probability $p(y_i|y_{<i},C)$ is computed in different ways according to the choice of the context $C$ at time $i$. In \cite{cho:2014}, the authors choose $C=\mathbf{h}^m_{T_x}$, while \newcite{bahdanau:2014} use different context $c_i$ at different time step and the conditional probability will become:

\begin{equation}
p(y_i|y_{<i},C) = p(y_i|y_{<i},c_i) = softmax(W\hat{z}_i))
\end{equation}

where $\hat{z}_i$ is the attention output:

\begin{equation}
\hat{z}_i = tahn(W_c[z^l_i;c_i])
\end{equation}

The attention model calculates $c_i$ as the weighted sum of the source-side context vectors, just as illustrated in the middle part of Fig. 2.

\begin{equation}
c_i = \sum_{j=1}^{T_x} \alpha_{ij}z^l_{i}
\end{equation}

where $\alpha_{ij}$ is a normalized item calculated as follows:

\begin{equation}
\alpha_{ij} = \frac{{\mathbf{h}^m_j} \cdot {z^l_i}}{{\sum_{j'} \mathbf{h}^m_{j'}} \cdot {z^l_i}}
\end{equation}

$z^k_i$ is computed using the following formula:

\begin{equation}
z^k_i = LSTM(z^{k}_{i-1}, z^{k-1}_i)
\end{equation}

If $k=1$, $z^1_i$ will be calculated by combining $\hat{z}_{i-1}$ as feed input \cite{luong:2015a}:

\begin{equation}
z^1_i = LSTM(z^1_{i-1}, y_{i-1}, \hat{z}_{i-1})
\end{equation}

Given the {\bf \em{sentence aligned bilingual training data}} $\mathcal{D}_b=\{(X_b^{(n)},Y_b^{(n)})\}_{n=1}^N$ , all the parameters of the encoder-decoder NMT are optimized to maximize the following conditional log-likelihood:

\begin{equation}
\mathcal{L}({\theta}) = \frac{1}{N} \sum_{n=1}^{N} \sum_{i=1}^{T_y} log p(y_i^{(n)}|y_{<i}^{(n)}, X^{(n)},\theta)
\end{equation}

\section{Incorporating Bilingual Dictionaries}

The word translation pairs in bilingual dictionaries are difficult to use in neural machine translation, mainly because they are rarely or never seen in the parallel training corpus. We attempt to build a bridge between NMT and bilingual dictionaries. We believe the bridge is {\bf {\em data transformation}} that can transform rarely or unseen word translation pairs into frequent ones and provide NMT adequate information to learn latent translation mappings. In this work, we propose two methods to perform data transformation from character level and word level respectively.

\subsection{Mixed Word/Character Model}

Given a {\bf \em{bilingual dictionary}} $\mathit{Dic}=\{(Dic_x^{(i)},Dic_y^{(i)})\}_{i=1}^I$, we focus on the translation lexicons $(Dic_x,Dic_y)$ if $Dic_x$ is a rare or unknown word in the bilingual corpus $\mathcal{D}_b$.

We first introduce data transformation using the character-based method. We all know that words are composed of characters and most of the characters are frequent even though the word is never seen. This idea is popularly used to deploy open vocabulary NMT \cite{ling:2015,costa:2016,chung:2016}.

Character translation mappings are much easier to learn for NMT than word translation mappings. However, given a character sequence of a source language word, NMT cannot guarantee the generated character sequence would lead to a valid target language word. Therefore, we prefer the framework mixing the words and characters, which is employed by \newcite{wu:2016} to handle OOV words. If it is a frequent word, we keep it unchanged. Otherwise, we fall back to the character sequence.

We perform data transformation on both parallel training corpus and bilingual dictionaries. Here, English sentences and words are adopted as examples. Suppose we keep the English vocabulary $V$ in which the frequency of each word exceeds a threshold $K$. For each English word $w$ (e.g. {\em oak}) in a parallel sentence pair $(X_b, Y_b)$ or in a translation lexicon $(Dic_x, Dic_y)$, if $w \in V$, $w$ will be left as it is. Otherwise, $w$ is re-labelled by character sequence. For example, {\em oak} will be:

\begin{equation}
oak \rightarrow \langle B \rangle o \quad \langle M \rangle a \quad \langle E \rangle k
\end{equation}

Where $\langle B \rangle$, $\langle M \rangle$ and $\langle E \rangle$ denotes respectively begin, middle and end of a word.

\subsection{Pseudo Sentence Pair Synthesis Model}

Since NMT is a data driven approach, it can learn latent translation mappings for a word pair $(Dic_x, Dic_y)$ if these exist many parallel sentences containing $(Dic_x, Dic_y)$. Along this line, we propose the pseudo sentence pair synthesis model. In this model, we aim at synthesizing for a rare or unknown translation lexicon $(Dic_x, Dic_y)$ the adequate pseudo parallel sentences $\{(X_p^j, Y_p^j)\}_{j=1}^J$ each of which contains $(Dic_x, Dic_y)$.

Although there are no enough bilingual sentence pairs in many languages (and many domains), a huge amount of the monolingual data is available in the web. In this paper, we plan to make use of the source-side monolingual data $\mathcal{D}_{sm}=\{(X_{sm}^{(m)}\}_{m=1}^M$ ($M \gg N$) to synthesize the pseudo bilingual sentence pairs $\mathcal{D}_{bp}=\{(X_p^j, Y_p^j)\}_{j=1}^J$.

\begin{algorithm}[t]
\caption{Pseudo Sentence Pair Synthesis.}
\label{alg:Framework}
\begin{algorithmic}[1]
\Require
	bilingual training data $\mathcal{D}_b$;
	bilingual dictionary $\mathit{Dic}$;
	source language monolingual data $\mathcal{D}_{sm}$;
	pseudo sentence pair number $K$ for each $(Dic_x,Dic_y)$;
\Ensure
	pseudo sentence pairs $\mathcal{D}_{bp}=\{(X_p^j, Y_p^j)\}_{j=1}^J$:
\State Build an SMT system {\bf \em PBMT} on $\{\mathcal{D}_b,\mathit{Dic}\}$;
\State $\mathcal{D}_{bp}=\{\}$;
\For {each $(Dic_x,Dic_y)$ in $\mathit{Dic}$}
\State Retrieve $K$ monolingual sentences $\{X_p^k\}_{k=1}^K$ containing $Dic_x$ from $\mathcal{D}_{sm}$;
\State Translate $\{X_p^k\}_{k=1}^K$ into target language sentences $\{Y_p^k\}_{k=1}^K$ using {\bf \em PBMT};
\State Add $\{X_p^k, Y_p^k\}_{k=1}^K$ into $\mathcal{D}_{bp}$;
\EndFor \\
\Return $\mathcal{D}_{bp}$
\end{algorithmic}
\end{algorithm}

For constructing $\mathcal{D}_{bp}$, we resort to statistical machine translation (SMT) and apply a {\em self-learning method} as illustrated in {\bf \em Algorithm 1}. In contrast to NMT, statistical machine translation (SMT, e.g. phrase-based SMT \cite{koehn:2007,xiong:2006}) is easy to integrate bilingual dictionaries \cite{wu:2008} as long as we consider the translation lexicons of bilingual dictionaries as phrasal translation rules. Following \cite{wu:2008}, we first merge the bilingual sentence corpus $\mathcal{D}_b$ with the bilingual dictionaries $\mathit{Dic}$, and employ the phrase-based SMT to train an SMT system called {\bf \em{PBMT}} (line 1 in {\em Algorithm 1}).

For each rare or unknown word translation pair $(Dic_x, Dic_y)$, we can easily retrieve the adequate source language monolingual sentences $\{(X_p^k)\}_{k=1}^K$ ($Dic_x \in X_p^k$) from the web or other data collections. {\bf \em{PBMT}} is then applied to translate $\{(X_p^k)\}_{k=1}^K$ to generate target language translations $\{(Y_p^k)\}_{k=1}^K$. As {\bf \em{PBMT}} employs the bilingual dictionaries $\mathit{Dic}$ as additional translation rules, each target translation sentence $Y_p \in \{(Y_p^k)\}_{k=1}^K$ will contain $Dic_y$. Then, the sentence pair $(X_p^k, Y_p^k)$ will include the word translation pair $(Dic_x, Dic_y)$. Finally, we can pair $\{(X_p^k)\}_{k=1}^K$ and $\{(Y_p^k)\}_{k=1}^K$ to yield pseudo sentence pairs $\{(X_p^k, Y_p^k)\}_{k=1}^K$, which will be added into $\mathcal{D}_{bp}$ (line 2-6 in {\em Algorithm 1}).

The original bilingual corpus $\mathcal{D}_b$ and the pseudo bilingual sentence pairs $\mathcal{D}_{bp}$ are combined together to train a new NMT model. Some may worry that the target parts of $\mathcal{D}_{bp}$ are SMT results but not well-formed sentences which would harm NMT training. Fortunately, \newcite{sennrich:2015a}, \newcite{cheng:2016b} and \newcite{zhang:2016} observe from large-scale experiments that the synthesized bilingual data using self-learning framework can substantially improve NMT performance. Since $\mathcal{D}_{bp}$ now contains bilingual dictionaries, we expect that the NMT trained on $\{\mathcal{D}_b, \mathcal{D}_{bp}\}$ cannot only significantly boost the translation quality, but also solve the problem of rare word translation if they are covered by $\mathit{Dic}$.

Note that the pseudo sentence pair synthesis model can be further augmented by the mixed word/character model to solve other OOV translations.

\section{Experimental Settings}

\begin{table*}\small
\begin{center}
\begin{tabular}{l|c|c|c|c|c|c|c|c}
\hline \bf Method & $|V_c|$ & $|V_e|$& \bf MT03 & \bf MT04 & \bf MT05 & \bf MT06 & \bf MT08 & \bf Ave \\ \hline
Moses & ~ & ~ & 30.30 & 31.04 & 28.19 & 30.04 & 23.20 & 28.55  \\
\hline
\hline
Zoph\_RNN & 38815 & 30514 & 34.77 & 37.40 & 32.94 & 33.85 & 25.93 & 32.98  \\ 
\hline
\hline
Zoph\_RNN-mixed & 42769 & 30630 & 35.57 & 38.07 & 34.44 & 36.07 & 26.81 & 34.19 \\
Zoph\_RNN-mixed-dic & 42892 & 30630 & 36.29 & 38.75 & 34.86 & 36.57 & 27.04 & 34.70 \\
\hline
\hline
Zoph\_RNN-pseudo ($K=10$) & 42133 & 32300 & 35.66 & 38.02 & 34.66 & 36.51 & 27.65 & 34.50 \\
Zoph\_RNN-pseudo-dic ($K=10$) & 42133 & 31734 & 36.48 & 38.59 & 35.81 & 38.14 & 28.65 & 35.53 \\
Zoph\_RNN-pseudo ($K=20$) & 43080 & 32813 & 35.00 & 36.99 & 34.22 & 36.09 & 26.80 & 33.82 \\
Zoph\_RNN-pseudo-dic ($K=20$) & 43080 & 32255 & 36.92 & 38.63 & 36.09 & 38.13 & 29.53 & 35.86 \\
Zoph\_RNN-pseudo ($K=30$) & 44162 & 33357 & 36.07 & 37.74 & 34.63 & 36.66 & 27.58 & 34.54 \\
Zoph\_RNN-pseudo-dic ($K=30$) & 44162 & 32797 & 37.26 & 39.01 & 36.64 & 38.50 & 30.17 & 36.32 \\
Zoph\_RNN-pseudo ($K=40$) & 45195 & 33961 & 35.44 & 37.96 & 34.89 & 36.92 & 27.80 & 34.60 \\
Zoph\_RNN-pseudo-dic ($K=40$) & 45195 & 33399 & 36.93 & 39.15 & 36.85 & 38.77 & 30.25 & 36.39 \\
\hline
\hline
Zoph\_RNN-pseudo-mixed ($K=40$) & 45436 & 32659 & 38.17 & 39.55 & 36.86 & 38.53 & 28.46 & 36.31 \\
Zoph\_RNN-pseudo--mixed-dic ($K=40$) & 45436 & 32421 & 38.66 & 40.78 & 38.36 & 39.56 & 30.64 & 37.60 \\
\hline
\end{tabular}
\caption{\label{Table 1} Translation results (BLEU score) for different translation methods. $K=10$ denotes that we synthesize 10 pseudo sentence pairs for each word translation pair $(Dic_x, Dic_y)$. The column $|V_c|$ ($|V_e|$) reports the vocabulary size limited by frequency threshold $u_c=10$ ($u_e=8$). Note that all the NMT systems use the single model rather than the ensemble model.}
\end{center}
\end{table*}

In this section we describe the data sets, data preprocessing, the training and evaluation details, and all the translation methods we compare in the experiments.

\subsection{Dataset}
We perform the experiments on Chinese-to-English translation. Our {\bf \em bilingual training data} $\mathcal{D}_b$ includes 630K{\footnote{Without using very large-scale data, it is relatively easy to evaluate the effectiveness of the bilingual dictionaries.}} sentence pairs (each sentence length is limited up to 50 words) extracted from LDC corpora{\footnote{LDC2000T50, LDC2002T01, LDC2002E18, LDC2003E07, LDC2003E14, LDC2003T17, LDC2004T07.}}. For validation, we choose NIST 2003 (MT03) dataset. For testing, we use NIST 2004 (MT04), NIST 2005 (MT05), NIST 2006 (MT06) and NIST 2006 (MT08) datasets. The test sentences are remained as their original length. As for the {\bf \em source-side monolingual data} $\mathcal{D}_{sm}$, we collect about 100M Chinese sentences in which approximately 40\% are provided by Sogou and the rest are collected by searching the words in the bilingual data from the web. We use two {\bf \em bilingual dictionaries}: one is from LDC (LDC2002L27) and the other is manually collected by ourselves. The combined dictionary $\mathit{Dic}$ contains 86,252 translation lexicons in total.

\subsection{Data Preprocessing}
If necessary, the Chinese sentences are word segmented using Stanford Word Segmenter{\footnote{http://nlp.stanford.edu/software/segmenter.shtml}}. The English sentences are tokenized using the tokenizer script from the Moses decoder{\footnote{http://www.statmt.org/moses/}}. We limit the vocabulary in both Chinese and English using a frequency threshold $u$. We choose $u_c=10$ for Chinese and $u_e=8$ for English, resulting $|V_c|=38815$ and $|V_e|=30514$ for Chinese and English respectively in $\mathcal{D}_b$. As we focus on rare or unseen translation lexicons of the bilingual dictionary $\mathit{Dic}$ in this work, we filter $\mathit{Dic}$ and retain the ones $(Dic_x, Dic_y)$ if $Dic_x \notin V_c$, resulting 8306 entries in which 2831 ones appear in the validation and test data sets. All the OOV words are replaced with \textit{UNK} in the word-based NMT and are re-labelled into character sequences in the mixed word/character model.

\subsection{Training and Evaluation Details}
We build the described models using the Zoph\_RNN{\footnote{https://github.com/isi-nlp/Zoph\_RNN}} toolkit which is written in C++/CUDA and provides efficient training across multiple GPUs. In the NMT architecture as illustrated in Fig. 2, the encoder includes two stacked LSTM layers, followed by a global attention layer, and the decoder also contains two stacked LSTM layers followed by the softmax layer. The word embedding dimension and the size of hidden layers are all set to 1000.

Each NMT model is trained on GPU K80 using stochastic gradient decent algorithm AdaGrad {\cite{duchi:2011}}. We use a mini batch size of $B=128$ and we run a total of 20 iterations for all the data sets. The training time for each model ranges from 2 days to 4 days. At test time, we employ beam search with beam size $b=10$. We use case-insensitive 4-gram BLEU score as the automatic metric \cite{papineni:2002} for translation quality evaluation.

\subsection{Translation Methods}
In the experiments, we compare our method with the conventional SMT model and the baseline attention-based NMT model. We list all the translation methods as follows:

\begin{itemize}
\item \textbf{Moses}: It is the state-of-the-art phrase-based SMT system {\cite{koehn:2007}}. We use its default configuration and train a 4-gram language model on the target portion of the bilingual training data.

\item \textbf{Zoph\_RNN}: It is the baseline attention-based NMT system {\cite{luong:2015b,zoph:2016a}} using two stacked LSTM layers for both of the encoder and the decoder.

\item \textbf{Zoph\_RNN-mixed-dic}: It is our NMT system which integrates the bilingual dictionaries by re-labelling the rare or unknown words with character sequence on both bilingual training data and bilingual dictionaries. {\em {Zoph\_RNN-mixed}} indicates that mixed word/character model is performed only on the bilingual training data and the bilingual dictionary is not used.

\item \textbf{Zoph\_RNN-pseudo-dic}: It is our NMT system that integrates the bilingual dictionaries by synthesizing adequate pseudo sentence pairs that contain the focused rare or unseen translation lexicons. {\em {Zoph\_RNN-pseudo}} means that the target language parts of pseudo sentence pairs are obtained by the SMT system {\bf \em{PBMT}} without using the bilingual dictionary $Dic$.

\item \textbf{Zoph\_RNN-pseudo-mixed-dic}: It is a NMT system combining the two methods {\em {Zoph\_RNN-pseudo}} and {\em {Zoph\_RNN-mixed}}. {\em {Zoph\_RNN-pseudo-mixed}} is similar to {\em {Zoph\_RNN-pseudo}}.
\end{itemize}

\section{Translation Results and Analysis}
For translation quality evaluation, we attempt to figure out the following three questions: 1) Could the employed attention-based NMT outperform SMT even on less than 1 million sentence pairs? 2) Which model is more effective for integrating the bilingual dictionaries: mixed word/character model or pseudo sentence pair synthesis data? 3) Can the combined two proposed methods further boost the translation performance?

\subsection{NMT vs. SMT}
Table 1 reports the detailed translation quality for different methods. Comparing the first two lines in Table 1, it is very obvious that the attention-based NMT system {\em Zoph\_RNN} substantially outperforms the phrase-based SMT system {\em Moses} on just 630K bilingual Chinese-English sentence pairs. The gap can be as large as 6.36 absolute BLEU points on MT04. The average improvement is up to 4.43 BLEU points (32.98 vs. 28.55). It is in line with the findings reported in \cite{wu:2016,junczys:2016} which conducted experiments on tens of millions or even more parallel sentence pairs. Our experiments further show that NMT can be still much better even we have less than 1 million sentence pairs.

\subsection{The Effect of The Mixed W/C Model}
The two lines (3-4 in Table 1) presents the BLEU scores when applying the mixed word/character model in NMT. We can see that this model markedly improves the translation quality over the baseline attention-based NMT, although the idea behind is very simple. 

Specifically, the system {\em Zoph\_RNN-mixed}, trained only on the bitext $\mathcal{D}_b$, achieves an average improvement of more than 1.0 BLEU point (34.19 vs 32.98) over the baseline {\em Zoph\_RNN}. It indicates that the mixed word/character model can alleviate the OOV translation problem to some extent. For example, the number {\em 31.3} is an OOV word in Chinese. The mixed model transforms this word into $\langle B\rangle 3 \quad \langle M\rangle 1 \quad \langle M\rangle . \quad \langle E\rangle 3$ and it is correctly copied into target side, yielding a correct translation {\em 31.3}. Moreover, some named entities (e.g. person name {\em hecker}) can be well translated.

When adding the bilingual dictionary $Dic$ as training data, the system {\em Zoph\_RNN-mixed-dic} further gets a moderate improvement of 0.51 BLEU points (34.70 vs 34.19) on average. We find that the mixed model could make use of some rare or unseen translation lexicons in NMT, as illustrated in the first two parts of Table 2. In the first part of Table 2, the English side of the translation lexicon is a frequent word (e.g. {\em remain}). The Chinese frequent character (e.g. $li\acute{u}$) shares the most meaning of the whole word ($zh\grave{u}li\acute{u}$) and thus it could be correctly translated into {\em remain}. We are a little surprised by the examples in the second part of Table 2, since the correct English parts are all OOV words which require each English character to be correctly generated. It demonstrates that the mixed model has some ability to predict the correct character sequence. However, this mixed model fails in many scenarios. The third part in Table 2 gives some bad cases. If the first predicted character is wrong, the final word translation will be incorrect (e.g. {{\em take-owned lane} vs. {\em overtaking lane}}). This is the main reason why the mixed model could not obtain large improvements.

\begin{table}\small
\begin{center}
\begin{tabular}{l|l|l}
\hline \bf Chinese Word & Translation & Correct \\ \hline
$zh\grave{u}li\acute{u}$ & remain & remain \\
$d\bar{o}ngji\bar{a}$ & owner & owner \\
$li\grave{e}y\grave{a}n$ & blaze & blaze \\
\hline
\hline
$\bar{a}nw\grave{e}ij\grave{i}$ & placebo & placebo \\
$h\check{a}ixi\grave{a}o$ & tsunami & tsunami \\
$j\grave{i}ngm\grave{a}i$ & intravenous & intravenous \\
\hline
\hline
$f\check{a}ny\grave{i}ngl\acute{u}$ & anti-subsidization & reactor \\
$hu\acute{a}ngp\check{u}ji\bar{a}ng$ & lingchiang river & huangpu river \\
$ch\bar{a}och\bar{e}d\grave{a}o$ & take-owned lane & overtaking lane \\
\hline
\end{tabular}
\caption{\label{Table 2} The effect of the {\em Zoph\_RNN-mixed-dic} model in using bilingual dictionaries. The Chinese word is written in Pinyin. The first two parts are positive word translation examples, while the third part shows some bad cases.}
\end{center}
\end{table}

\subsection{The Effect of Data Synthesis Model}
The eight lines (5-12) in Table 1 show the translation performance of the pseudo sentence pair synthesis model. We can analyze the results from three perspectives: 1) the effect of the self-learning method for using the source-side monolingual data; 2) the effect of the bilingual dictionary; and 3) the effect of pseudo sentence pair number.

The results in the odd lines (lines with {\em Zoph\_RNN-pseudo}) demonstrate that the synthesized parallel sentence pairs using source-side monolingual data can significantly improve the baseline NMT {\em Zoph\_RNN}, and the average improvement can be up to 1.62 BLEU points (34.60 vs. 32.98). This finding is also reported by \newcite{cheng:2016b} and \newcite{zhang:2016}.

After augmenting {\em Zoph\_RNN-pseudo} with bilingual dictionaries, we can further obtain considerable gains. The largest average improvement can be 3.41 BLEU points when compared to the baseline NMT {\em Zoph\_RNN} and 2.04 BLEU points when compared to {\em Zoph\_RNN-pseudo} (35.86 vs. 33.82).

When investigating the effect of pseudo sentence pair number (from $K=10$ to $K=40$), we find that the performance is largely better and better if we synthesize more pseudo sentence pairs for each rare or unseen word translation pair $(Dic_x, Dic_y)$. We can also notice that improvement gets smaller and smaller when $K$ grows.

\begin{table}\small
\begin{center}
\begin{tabular}{c|c|c|c|c}
\hline
& \multicolumn{2}{|c}{\centering pseudo-dic}\\
\cline{2-5}
\multicolumn{1}{c|}{mixed-dic}
& \multicolumn{1}{c|}{$K=10$}
& \multicolumn{1}{c|}{$K=20$}
& \multicolumn{1}{c|}{$K=30$}
& \multicolumn{1}{c}{$K=40$}\\
\hline
0.36 & 0.71 & 0.76 & 0.78 & 0.79\\
\hline
\end{tabular}
\caption{\label{Table 3} The hit rate of the bilingual dictionary for different models.}
\end{center}
\end{table}

\subsection{Mixed W/C Model vs. Data Synthesis Model}
Comparing the results between the mixed model and the data synthesis model ({\em Zoph\_RNN-mixed-dic} vs. {\em Zoph\_RNN-pseudo-dic}) in Table 1, we can easily see that the data synthesis model is much better to integrate bilingual dictionaries in NMT. {\em Zoph\_RNN-pseudo-dic} can substantially outperform {\em Zoph\_RNN-mixed-dic} by an average improvement up to 1.69 BLEU points (36.39 vs. 34.70).

Through a deep analysis, we find that most of rare or unseen words in test sets can be well translated by {\em Zoph\_RNN-pseudo-dic} if they are covered by the bilingual dictionary. Table 3 reports the hit rate of the bilingual dictionaries. $0.71$ indicates that $2010$ ($2831 \times 0.71$) words among the $2831$ covered rare or unseen words in the test set can be correctly translated. This table explains why {\em Zoph\_RNN-pseudo-dic} performs much better than {\em Zoph\_RNN-mixed-dic}.

The last two lines in Table 1 demonstrate that the combined method can further boost the translation quality. The biggest average improvement over the baseline NMT {\em Zoph\_RNN} can be as large as 4.62 BLEU points, which is very promising. We believe that this method fully exploits the capacity of the data synthesis model and the mixed model. {\em Zoph\_RNN-pseudo-dic} can well incorporate the bilingual dictionary and {\em Zoph\_RNN-mixed} can well handle the OOV word translation. Thus, the combined method is the best.

One may argue that the proposed methods use bigger vocabulary and the performance gains may be attributed to the increased vocabulary size. We further conduct an experiment for the baseline NMT {\em Zoph\_RNN} by setting $|V_c|=4600$ and $|V_e|=3400$. We find that this setting decreases the translation quality by an average BLEU points 0.88 (32.10 vs. 32.98). This further verifies the superiority of our proposed methods.

\section{Related Work}

The recently proposed neural machine translation has drawn more and more attention. Most of the existing methods mainly focus on designing better attention models {\cite{luong:2015a,cheng:2016a,cohn:2016,feng:2016,liu:2016,meng:2016,mi:2016a,mi:2016b,tu:2016}}, better objective functions for BLEU evaluation {\cite{shen:2016}}, better strategies for handling open vocabulary {\cite{ling:2015,luong:2015c,jean:2015,sennrich:2015a,costa:2016,lee:2016,li:2016oov,mi:2016c,wu:2016}} and exploiting large-scale monolingual data {\cite{gulcehre:2015,sennrich:2015b,cheng:2016b,zhang:2016}.

Our focus in this work is aiming to fully integrate the discrete bilingual dictionaries into NMT. The most related works lie in three aspects: 1) applying the character-based method to deal with open vocabulary; 2) making use of the synthesized data in NMT, and 3) incorporating translation lexicons in NMT.

{\newcite{ling:2015}}, {\newcite{costa:2016}} and {\newcite{sennrich:2015a}} propose purely character-based or subword-based neural machine translation to circumvent the open word vocabulary problem. {\newcite{luong:2015c}} and {\newcite{wu:2016}} present the mixed word/character model which utilizes character sequence to replace the OOV words. We introduce the mixed model to integrate the bilingual dictionaries and find that it is useful but not the best method.

{\newcite{sennrich:2015b}} propose an approach to use target-side monolingual data to synthesize the bitexts. They generate the synthetic bilingual data by translating the target monolingual sentences to source language sentences and retrain NMT with the mixture of original bilingual data and the synthetic parallel data. {\newcite{cheng:2016b}} and {\newcite{zhang:2016}} also investigate the effect of the synthesized parallel sentences. They report that the pseudo sentence pairs synthesized using the source-side monolingual data can significantly improve the translation quality. These studies inspire us to leverage the synthesized data to incorporate the bilingual dictionaries in NMT.

Very recently, \newcite{arthur:2016} try to use discrete translation lexicons in NMT. Their approach attempts to employ the discrete translation lexicons to positively influence the probability distribution of the output words in the NMT softmax layer. However, their approach only focuses on the words that belong to the vocabulary and the out-of-vocabulary (OOV) words are not considered. In contrast, we concentrated ourselves on the word translation lexicons which are rarely or never seen in the bilingual training data. It is a much tougher problem. The extensive experiments demonstrate that our proposed models, especially the data synthesis model, can solve this problem very well.

\section{Conclusions and Future Work}
In this paper, we have presented two models to bridge neural machine translation and the bilingual dictionaries in which translation lexicons are rarely or never seen in the bilingual training data. Our proposed methods focus on data transformation mechanism which guarantees the massive and repetitive occurrence of the translation lexicon.

The mixed word/character model tackles this problem by re-labelling the OOV words with character sequence, while our data synthesis model constructs adequate pseudo sentence pairs for each translation lexicon. The extensive experiments show that the data synthesis model substantially outperforms the mixed word/character model, and the combined method performs best. All of the proposed methods obtain promising improvements over the baseline NMT. We further find that more than 70\% of the rare or unseen words in test sets can get correct translations as long as they are covered by the bilingual dictionary.

Currently, the data synthesis model does not distinguish the original bilingual training data from the synthesized parallel sentences in which the target sides are SMT translation results. In the future work, we plan to modify the neural network structure to avoid the negative effect of the SMT translation noise.


\bibliography{acl2016}
\bibliographystyle{acl2016}


\end{document}